\documentclass{article}

\usepackage{todonotes}
\usepackage{tikz}

\usetikzlibrary{arrows.meta, bending, decorations.pathreplacing, bayesnet}

 \usepackage[preprint]{neurips_2025}

\workshoptitle{Principles of Generative Modeling (PriGM)}

\usepackage[utf8]{inputenc} %
\usepackage[T1]{fontenc}    %
\usepackage{hyperref}       %
\usepackage{url}            %
\usepackage{booktabs}       %
\usepackage{amsfonts}       %
\usepackage{nicefrac}       %
\usepackage{microtype}      %
\usepackage{xcolor}         %
\usepackage{todonotes}
\title{Masked Diffusion Models are Secretly Learned-Order Autoregressive Models}

\author{%
  Prateek Garg$^*$ \\
  Aalto University \\
  \And
  Bhavya Kohli\thanks{Work done while at IIT Bombay \\ \phantom{sfsfs}Correspondence to: Prateek Garg \texttt{<prateekg@iitb.ac.in>}} \\
  NUS, Singapore \\
  \And
  Sunita Sarawagi \\
  IIT Bombay \\
}

\usepackage{amsmath,amsfonts,bm}

\def\eqref#1{equation~\ref{#1}}

\def\1{\bm{1}}

\DeclareMathAlphabet{\mathsfit}{\encodingdefault}{\sfdefault}{m}{sl}
\SetMathAlphabet{\mathsfit}{bold}{\encodingdefault}{\sfdefault}{bx}{n}

\DeclareMathOperator*{\argmax}{arg\,max}

\def\x{{\mathbf x}}

\def\t{{t(i)}}
\def\bmu{{\bm \mu}}

\def\bmu{{\bm \mu}}

\def\lossfinal{{\mathcal{L} (\x_0)}}
\def\lossfinalseq{{\mathcal{L} (\x_0^{1:L})}}

\def\x{{\mathbf x}}

\def\dt{{\text{d}}t}

\def\m{{\mathbf m}}

\def\t{{t(i)}}
\def\bmu{{\bm \mu}}

\def\bmu{{\bm \mu}}

\def\lossfinal{{\mathcal{L} (\x_0)}}

\def\ats{\alpha_{t|s}}
\def\bts{\beta_{t|s}}

\def\at{\alpha_{t}}
\def\ati{\alpha_{t(i)}}
\def\asi{\alpha_{s(i)}}
\def\atj{\alpha_{t(j)}}
\def\asj{\alpha_{s(j)}}
\def\dat{\alpha'_{t}}

\def\atl{\alpha_{t, \ell}}
\def\datl{\alpha'_{t, \ell}}

\def\as{\alpha_{s}}
\def\denoise{\mathbf {\mu}_\theta}

\usepackage{prettyref}
\newcommand{\pref}[1]{\prettyref{#1}}
\newcommand{\pfref}[1]{Proof of \prettyref{#1}}
\newcommand{\savehyperref}[2]{\texorpdfstring{\hyperref[#1]{#2}}{#2}}
\newrefformat{eq}{\savehyperref{#1}{\textup{(\ref*{#1})}}}
\newrefformat{eqn}{\savehyperref{#1}{Equation~\ref*{#1}}}
\newrefformat{lem}{\savehyperref{#1}{Lemma~\ref*{#1}}}
\newrefformat{def}{\savehyperref{#1}{Definition~\ref*{#1}}}
\newrefformat{line}{\savehyperref{#1}{line~\ref*{#1}}}
\newrefformat{thm}{\savehyperref{#1}{Theorem~\ref*{#1}}}
\newrefformat{corr}{\savehyperref{#1}{Corollary~\ref*{#1}}}
\newrefformat{sec}{\savehyperref{#1}{Section~\ref*{#1}}}
\newrefformat{ssec}{\savehyperref{#1}{Subsection~\ref*{#1}}}
\newrefformat{app}{\savehyperref{#1}{Appendix~\ref*{#1}}}
\newrefformat{ass}{\savehyperref{#1}{Assumption~\ref*{#1}}}
\newrefformat{ex}{\savehyperref{#1}{Example~\ref*{#1}}}
\newrefformat{fig}{\savehyperref{#1}{Figure~\ref*{#1}}}
\newrefformat{alg}{\savehyperref{#1}{Algorithm~\ref*{#1}}}
\newrefformat{rem}{\savehyperref{#1}{Remark~\ref*{#1}}}
\newrefformat{conj}{\savehyperref{#1}{Conjecture~\ref*{#1}}}
\newrefformat{prop}{\savehyperref{#1}{Proposition~\ref*{#1}}}
\newrefformat{proto}{\savehyperref{#1}{Protocol~\ref*{#1}}}
\newrefformat{prob}{\savehyperref{#1}{Problem~\ref*{#1}}}
\newrefformat{claim}{\savehyperref{#1}{Claim~\ref*{#1}}}
\newrefformat{fact}{\savehyperref{#1}{Fact~\ref*{#1}}}

\usepackage{nicefrac}
\usepackage{amsthm}
\usepackage{amssymb}
\usepackage{hyperref}
\hypersetup{
    colorlinks,
    linkcolor={red!50!black},
    citecolor={blue!50!black},
    urlcolor={blue!50!black}
}
\usepackage{graphicx}
\usepackage{subfig}
\usepackage{wrapfig}
\usepackage{url}
\usepackage{algorithm}
\usepackage{algpseudocode}

\usepackage{thm-restate}

\newtheoremstyle{myplain}%
{3pt}%
{3pt}%
{\itshape}%
{}%
{\bfseries}%
{}%
{.5em}%
{}%

\newtheoremstyle{mypro}%
{3pt}%
{3pt}%
{}%
{}%
{\bfseries}%
{}%
{.5em}%
{}%
\theoremstyle{myplain}
\newtheorem{theorem}{Theorem}[section]
\newtheorem{corollary}[theorem]{Corollary}

\theoremstyle{mypro}

\newtheorem*{remark*}{Remark} %

\theoremstyle{definition}

\usepackage{macros}
\usepackage{multirow}

\begin{document}

\maketitle

\begin{abstract}
Masked Diffusion Models (MDMs) have emerged as one of the most promising paradigms for generative modeling over discrete domains. It is known that MDMs effectively train to decode tokens in a uniformly random order, and that this ordering has significant performance implications in practice. 
This observation raises a fundamental question: can we design a training framework that optimizes for a favorable decoding order?
We answer this in the affirmative, showing that the continuous-time variational objective of MDMs, when equipped with multivariate noise schedules, can identify and optimize for a decoding order during training. 
We establish a direct correspondence between decoding order and the multivariate noise schedule and show that this setting breaks invariance of the MDM objective to the noise schedule. 
Furthermore, we prove that the MDM objective decomposes precisely into a weighted auto-regressive losses over these orders, which establishes them as auto-regressive models with learnable orders.
\end{abstract}

\section{Introduction}
Autoregressive models (ARMs) remain the dominant paradigm for sequential data generation, largely due to their natural alignment with next-token prediction tasks on domains such as text. While text data possesses an inherent left-to-right sequential structure, many other discrete modalities of interest lack such canonical orderings. 
For example, tabular data~\citep{zhu_permutation-invariant_2022}, graphs~\citep{bu_let_2023}, 1D tokenized images~\citep{yu_image_2024} present ambiguity in how dimensions should be ordered for ARMs. 
This limitation of ARMs has also motivated research in model classes based on diffusion~\citep{sohl-dickstein15, ho2022classifierfreediffusionguidance, dhariwal2021diffusion, ddpm, ddim, vdm, edm}. 
While diffusion models have demonstrated remarkable success in continuous domains--including image synthesis, video generation, and speech processing--their application to discrete structures remains an active research area~\citep{d3pm,shi_simplified_2025, yu_discrete_2025, gat_discrete_2024, lou_discrete_2024, rutte_generalized_2025, sahoo_diffusion_2025}. 
Extending diffusion-like dynamics to discrete data modalities such as natural language, molecular structures, and protein sequences is a compelling alternative to ARMs which can potentially enable parallel generation and provide trade-offs between quality and sampling speed. 
Recent research~\citep{sahoo_simple_2024, shi_simplified_2025} has positioned one variant of discrete diffusion, Masked Diffusion Models (MDMs) as particularly promising, by extending their formulations to continuous time, combining the theoretical grounding of diffusion processes with the simplicity of masked token modeling.

Recent theoretical analyses \citep{zheng_masked_2025, ou_your_2025} have revealed a fundamental equivalence: the learning objective of MDMs corresponds exactly to masked language modeling and can be interpreted as any-order autoregressive models (AO-ARMs)~\citep{hoogeboom_autoregressive_2022}. 
In other words, the MDM training objective optimizes equally over all possible orders.
\citet{kim_train_2025} introduces inference-time heuristics to select a state-dependent ordering which achieves improvements on reasoning tasks such as sudoku. 
This observation suggests that MDMs possess the capacity to discover and exploit ordering structures during training, with certain orderings proving more advantageous than others even when the training objective weighs each ordering equally. 
This is an instance of train-test mismatch, while training weighs all the ordering equally, inference does not. 
This motivates us to extend the MDM objective, which can discover the order during training and thus optimize for that order. 
Concurrently, recent work~\citep{wang_learning-order_2025} investigates Learned-Order Autoregressive Models~(LO-ARMs) which attempts to learn a state-dependent order distribution. We instead pose a simpler question: \textit{Can the masked diffusion training objective be extended in a principled manner to learn optimal state-independent orderings?}

\paragraph{Our Contributions} We prove that when extended with multivariate noise schedules, the masked diffusion objective decomposes exactly as the expectation over possible orderings~(\pref{prop:elbo-lo-arm}), where the probability of sampling orders is defined by the noise schedule. 
We establish an exact correspondence of token ordering and multivariate noise schedule at inference time~(\pref{prop:cdf-transition-time}). 
In \pref{sec:expts}, we validate our theoretical claims with experimental results on tabular generative modeling.%

\section{Background}\vspace{-0.5em}
\begin{figure}[h]
    \centering
    \caption{Forward process of masked diffusion, masks variables in a order. While for univariate noise schedules, this order is uniformly random, multivariate noise schedule makes some order more likely than others.}
    \includegraphics[width=1\linewidth]{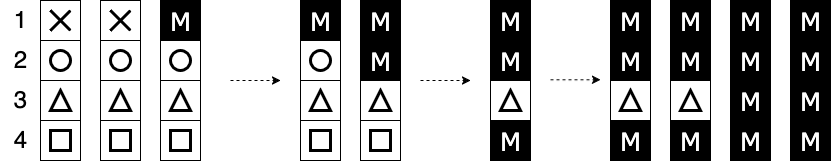}
    \label{fig:forwardprocps}
\end{figure}\vspace{-6pt}
Masked diffusion assumes a forward process where a token $\x_0$ transitions to a special mask state $\m$ at a random time defined by a noise schedule $\at$. For time $t\geq s$,
\begin{align}\label{eqn:interpolating_forward}
    q(\x_t | \x_s) = \ats \delta_{\x_s} + (1 - \ats)\delta_\m
\end{align}
where $\alpha_t \in [0, 1]$ is a strictly decreasing function in $t$, with $\alpha_{0} \approx 1$, $\alpha_{1} \approx 0$, and $\ats = \at/\as$.
The forward process can be interpreted as follows: Between timestep $s$ and $t$, we transition to a masked state with probability $1 - \ats$.
Once masked, a token stays masked with probability $1$.
This forward process admits a closed form posterior. 
For $s<t$, 
\begin{align}\label{eqn:true_posterior}
    q(\x_s | \x_t, \x_0) =
    \begin{cases}
        \delta_{\x_t} & \x_t \neq \m, \\
        \frac{1 - \as}{1- \at}\delta_\m + \frac{\as - \at}{1- \at}\delta_{\x_0}  & \x_t=\m. 
    \end{cases}
\end{align}
The reverse process conditioned on $\x_0$ also has a simple interpretation--if $\x_t$ is masked, it will jump to the state $\x_0$ at time $s$ with probability $\frac{\as - \at}{1 - \at}$, and stay masked otherwise. 
Once $\x_t$ is unmasked, it remains unchanged until $t=0$.
To learn a reverse model $p_\theta$, we optimize a variational bound (ELBO) on likelihoods. 
For a given a number of discretization steps $T$, we define $s(i) := (i -1)/T$ and $t(i) := i /T$, then we can write the discrete time ELBO as:
\begin{align}\label{eqn:elbo-trajectory}
    - \log p_\theta(\x_0)&\leq\mathbb{E}_q\!\left[
    - \log p(\x_T)
    - \sum_{i \ge 1}
        \log \frac{p_\theta(\x_{s(i)} \mid \x_{t(i)})}
                   {q(\x_{t(i)}\mid \x_{s(i)})}
\right] = \lossfinal
\end{align}

Under mean-parameterization~\citep{mdlm, shi_simplified_2025}, the reverse model learns a distribution over $\x_0$ given $\x_t$; $\x_0 \sim \denoise(\x_t,t ;\theta)$ and plugs into the closed form expression~(\ref{eqn:true_posterior}). 
For a sequence $\x^{1:L}$ consisting of $L$ tokens, such a reverse model can be optimized with the continuous time loss:
\begin{align}\label{eqn:difflossseqqqq}
    \lossfinalseq
    = \int_{t=0}^{t=1} \frac{\dat}{1 - \at}\left(\sum_{\ell} \mathbb{E}_{\x_t \sim q(\cdot | \x_0)}[\log \bmu^{\x^\ell_0}_\ell(\x_t,t ; \theta)] \right)\dt
\end{align}

where $\denoise^{\x_0}$ refers to the probability corresponding to the specific token $\x_0$ from a distribution defined over the space of all possible values of $\x_0$.

\section{Equivalence of Masked Diffusion and Learned-Order Autoregression}\vspace{-3pt}
In this section, we will explore the behavior of masked diffusion when each position has a different noise schedule $\atl$, which we refer to as a multivariate noise schedule.\vspace{-5pt}

\subsection{Reframing ELBO as Loss over Orders}

\begin{restatable}{proposition}{elboloarm}\label{prop:elbo-lo-arm}
    The diffusion loss (\pref{eqn:elbo-trajectory}) can be decomposed over the orders as follows:
    \begin{align}
        \lossfinalseq &= -\mathbb{E}_{\pi}\left[\sum_i\mathbb{E}_{t^*_{\pi(i)}|\pi}\left[\log\bmu(\x_{\pi(i)}|\x_{\pi(<i)},t^*_{\pi(i)}; \theta)\right]\right] 
    \end{align}
where $\bmu(\x_{\pi(i)}|\x_{\pi(<i)},t^*_{\pi(i)};\theta) = \bmu^{\x^{\pi(i)}_0}_{\pi(i)}(\x_{\pi(<i)}, t^*_{\pi(i)} ; \theta)$, $\x_{\pi(<i)}$ is a sequence obtained by masking the $\pi(\geq i)$ indices of $\x_0$ and $t^*_{\pi(i)}$ is the transition time of position $\pi(i)$. Under a time-independent network parameterization, 
    \begin{align}
        \lossfinalseq &= -\mathbb{E}_{\pi}\left[\sum_i \log \bmu(\x_{\pi(i)}|\x_{\pi(<i)}; \theta)\right] = -\mathbb{E}_{\pi}\left[LL_\pi(\x_0, \theta)\right]
    \end{align}
\end{restatable}
where $LL_\pi$ is the auto-regressive log-likelihood computed under order $\pi$ and expectation is taken over all orders $\pi$, where the corresponding probabilities are given as
 \begin{align}\label{eqn:order-distribution}
    P(\pi) = \int_{\Omega_\pi} \left(\prod_\ell -\datl\right) 
\end{align}
where $\Omega_\pi = \{t: t_{\pi(1)} < t_{\pi(2)} < \cdots < t_{\pi(L)} \}$

This expression can be interpreted as Learning-Order Autoregressive Models~\citep{wang_learning-order_2025}, with the key difference that our distribution over orders is state-independent (\pref{eqn:order-distribution})
\subsection{Relationship between Decoding Order and Noise Schedules}
\begin{restatable}{proposition}{cdftranstime}\label{prop:cdf-transition-time}
    Let $t^*$ be the time left in the reverse process with schedule $\at$ when a token transitions to a non-mask state, then $P(t^* \leq t) = 1 - \at$
\end{restatable}
\begin{corollary}
    For the $\ell$-th variable, the transition time $t^*_\ell$ is distributed according to the p.d.f $-\datl$
\end{corollary}

This gives us a way to sample an order $\pi$ given the schedule. 
We have outlined the sampling process in~\pref{alg:order-sampling}. 
\begin{algorithm}[H]
\caption{Order Sampler}
\label{alg:order-sampling}
\begin{algorithmic}[1]
\Require Noise schedules $\{\atl\}_{\ell=1}^L$
\Ensure Sampled order $\pi$
\State Initialize an empty list $\mathcal{T} \gets \varnothing$.
\For{$\ell = 1$ to $L$}
    \State Sample $u \sim \mathrm{Uniform}(0,1)$
    \State Compute $t_\ell \gets \alpha_\ell^{-1}(u)$ \Comment{sample $t_i$ }
    \State Append $t_\ell$ to $\mathcal{T}$
\EndFor
\State $\pi \gets $ \textsc{Sorted}$(\mathcal{T})$
\State \Return $\pi$
\end{algorithmic}
\end{algorithm}

This algorithm when used with univariate schedule also simplifies FHS sampler proposed by~\cite{zheng_masked_2025} and ``diffusion denoising
schedule'' used in ~\cite{sahoo_esoteric_2025} for univariate schedules.
\section{Experiments}\label{sec:expts}
In this section, we apply a masked diffusion with feature wise schedules to the task of tabular data generation. The forward process is applied independently across the sequence but  with different noise schedules $\atl$ across the sequence. The loss we use is:
\begin{align}\label{eqn:difflossseq}
    \lossfinalseq
    = \int_{t=0}^{t=1} \sum_{\ell} \frac{\datl}{1 - \atl} \mathbb{E}_{\x_t \sim q(\cdot | \x_0)}[\log \bmu^{\x^\ell_0}_\ell(\x_t; \theta)] \dt
\end{align}
Note the difference with~\pref{eqn:difflossseqqqq}, each dimension is weight defined by corresponding schedules. We parameterize $\atl = 1 - t^{w_\ell}$. To differentiate through the masking process we use RLOO gradient estimation by ~\citet{kool_buy_2019}, denoted by \rloosmall, and compare it with a masked diffusion model trained with fixed linear schedule (denoted \normalsmall).\vspace{-5pt}

\subsection{Results}\vspace{-5pt}
In this section, and in~\pref{app:addperf}, we compare the performance of well established tabular synthesis baselines on several data fidelity metrics. 
We refer readers to~\citet{shi_tabdiff_2025} for descriptions of the baselines and metrics used. 
For all the datasets, we implement \normalsmall~and \rloosmall~using the same hyperparameter set. All models trained by us have around $86$K parameters. We also note that many of the baseline methods make use of gaussian diffusion which typically requires a much higher number of steps than MDMs, whereas in our case the max number of steps is simply the number of columns. 

In Tables \ref{tab:trend}-\ref{tab:alphabeta}, \underline{Underlined numbers} denote the best performing method for each dataset. In Table \ref{tab:trend}, specifically for the Trend metric, we see that while not the best, both \normalsmall~and \rloosmall~are very competitive with state of the art baselines while having a fraction of the parameters--for example, TabDiff utilises models with $10$M parameters. We note similar performance comparisons on the other metrics. We also report additional metrics with standard deviations across runs in~\pref{app:addperf}.
\vspace{-1em}

\begin{table}[htbp]
\centering\caption{Performance comparison of our models on \textbf{\densitytrend}~against baselines across six datasets. The \densitytrend~metric measures the quality of pairwise correlations.}\label{tab:trend}\vspace{6pt}
\begin{tabular}{lrrrrrr}
\toprule
Method & \ccol{Adult} & \ccol{Default} & \ccol{Beijing} & \ccol{Shoppers} & \ccol{Magic} & \ccol{News} \\
\midrule
STaSy & $0.8549$ & $0.9404$ & $0.9151$ & $0.9339$ & $0.9200$ & $0.9693$ \\
CoDi & $0.7751$ & $0.3159$ & $0.8222$ & $0.9347$ & $0.9293$ & $0.8890$ \\
TabDDPM & $0.9699$ & $0.9511$ & $0.9339$ & $0.9830$ & $0.9729$ & $0.8684$ \\
TabSYN & $0.9807$ & $0.9719$ & $0.9787$ & $0.9912$ & $0.9687$ & $0.9848$ \\
TabDIFF & $\tablehighlight{0.9851}$ & $\tablehighlight{0.9745}$ & $\tablehighlight{0.9826}$ & $\tablehighlight{0.9924}$ & $\tablehighlight{0.9741}$ & $\tablehighlight{0.9872}$ \\
\midrule\rloosmall & $0.9778$ & $0.9705$ & $0.9743$ & $0.9710$ & $0.9628$ & $0.9781$ \\
\normalsmall & $0.9747$ & $0.9642$ & $0.9773$ & $0.9703$ & $0.9724$ & $0.9766$ \\
\bottomrule
\end{tabular}
\end{table}%

\begin{table}[H]
\centering\caption{Performance comparison of our models on \textbf{\densityshape}~against baselines across six datasets. The \densityshape~metric measures how well synthetic data captures each column’s marginal density.}\label{tab:shape}\vspace{6pt}
\begin{tabular}{lrrrrrr}
\toprule
Method & \ccol{Adult} & \ccol{Default} & \ccol{Beijing} & \ccol{Shoppers} & \ccol{Magic} & \ccol{News} \\
\midrule
STaSy & $0.8871$ & $0.9423$ & $0.9063$ & $0.9371$ & $0.9329$ & $0.9311$ \\
CoDi & $0.7862$ & $0.8423$ & $0.6816$ & $0.8844$ & $0.8306$ & $0.6773$ \\
TabDDPM & $0.9825$ & $0.9843$ & $0.9728$ & $0.9899$ & $0.9870$ & $0.2125$ \\
TabSyn & $0.9919$ & $\tablehighlight{0.9899}$ & $0.9856$ & $0.9897$ & $0.9874$ & $\tablehighlight{0.9794}$ \\
TabDiff & $\tablehighlight{0.9937}$ & $0.9876$ & $\tablehighlight{0.9872}$ & $\tablehighlight{0.9922}$ & $\tablehighlight{0.9897}$ & $0.9765$ \\
\midrule\rloosmall & $0.9878$ & $0.9830$ & $0.9841$ & $0.9832$ & $0.9750$ & $0.9787$ \\
\normalsmall & $0.9851$ & $0.9787$ & $0.9866$ & $0.9829$ & $0.9822$ & $0.9785$ \\
\bottomrule
\end{tabular}
\end{table}

\begin{figure}[H]
    \centering
    \caption{Comparing the best validation losses for \rloosmall~and \normalsmall~and visualizing the noise schedules learned by \rloosmall~on the Adult dataset.}
    
    \subfloat[Validation performance]{\label{tab:valloss}
        \parbox[t][2cm][b]{.45\textwidth}{
        \centering
        \begin{tabular}{lcc}
        \toprule
        Dataset  & \ccol{\rloosmall}          & \ccol{\normalsmall}        \\\midrule
        Adult    & $\tablehighlight{13.1524}$ & $13.1861$                  \\
        Default  & $\tablehighlight{18.3272}$ & $19.4247$                  \\
        Beijing  & $17.0060$                  & $\tablehighlight{16.7610}$ \\
        Shoppers & $16.4604$                  & $\tablehighlight{16.4246}$ \\
        Magic    & $\tablehighlight{12.4712}$ & $12.6114$                  \\
        News     & $\tablehighlight{48.9401}$ & $49.9699$                  \\
        \bottomrule
        \end{tabular}
        }
    }
    \subfloat[Learned noise schedules on Adult Dataset]{\label{fig:learnedschedules}
        \parbox[t][2cm][t]{.45\textwidth}{
        \centering \includegraphics[width=0.4\textwidth]{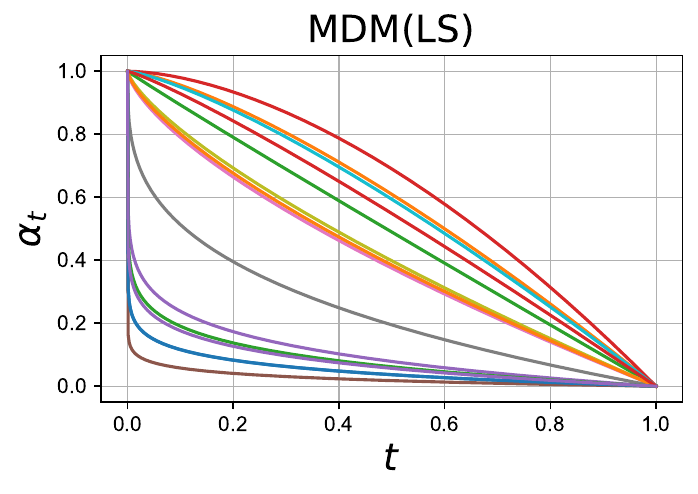}
    }}
\end{figure}\vspace{-2em}

From Table~\ref{tab:valloss}, we see that \rloosmall~leads to lower validation losses, although the performance gains are limited, as can be seen from Table~\ref{tab:trend}. Among the baselines, TabDiff~\citep{shi_tabdiff_2025} also utilises learned schedules.
In~\pref{fig:learnedschedules}, we visualize the schedules learned by \rloosmall~for the Adult dataset (Other datasets are provided in~\pref{fig:noise_md4}). In contrast, TabDiff fails to learn any schedules as seen in~\pref{app:learnedschedule}.\vspace{-7pt}

\begin{table}[t]
\centering
\caption{Performance comparison of our models on the data fidelity metrics \alphaprecision~and \betarecall, across the six datasets.}\label{tab:alphabeta}\vspace{6pt}
\resizebox{\textwidth}{!}{
\begin{tabular}{l
cccccc cccccc}
\toprule
\multirow{2}{3em}{\vspace{-5pt}Method} & \multicolumn{6}{c}{\alphaprecision}
& \multicolumn{6}{c}{\betarecall} \\
\cmidrule(lr){2-7}\cmidrule(lr){8-13}
 &
\ccol{Adult} & \ccol{Default} & \ccol{Beijing} & \ccol{Shoppers} & \ccol{Magic} & \ccol{News} &
\ccol{Adult} & \ccol{Default} & \ccol{Beijing} & \ccol{Shoppers} & \ccol{Magic} & \ccol{News} \\
\midrule

STaSy
& $0.8287$ & $0.9048$ & $0.8965$ & $0.8656$ & $0.8916$ & $0.9476$
& $0.2921$ & $0.3931$ & $0.3724$ & $\tablehighlight{0.5397}$ & $0.5479$ & $0.3942$ \\

CoDi
& $0.7758$ & $0.8238$ & $0.9495$ & $0.8501$ & $\tablehighlight{0.9813}$ & $0.8715$
& $0.0920$ & $0.1994$ & $0.2082$ & $0.5056$ & $0.5219$ & $0.3440$ \\

TabDDPM
& $0.9636$ & $0.9759$ & $0.8855$ & $0.9859$ & $0.9793$ & $0.0000$
& $0.4705$ & $0.4783$ & $0.4779$ & $0.4846$ & $0.5692$ & $0.0000$ \\

TabSyn
& $\tablehighlight{0.9939}$ & $\tablehighlight{0.9865}$ & $0.9836$ & $0.9942$ & $0.9751$ & $0.9505$
& $0.4792$ & $0.4645$ & $0.4910$ & $0.4803$ & $0.5915$ & $\tablehighlight{0.4301}$ \\

TabDiff
& $0.9902$ & $0.9849$ & $\tablehighlight{0.9911}$ & $\tablehighlight{0.9947}$ & $0.9806$ & $0.9736$
& $\tablehighlight{0.5164}$ & $\tablehighlight{0.5109}$ & $0.4975$ & $0.4801$ & $\tablehighlight{0.5963}$ & $0.4210$ \\\midrule

\rloosmall
& $0.9909$ & $0.9839$ & $0.9744$ & $0.9818$ & $0.9623$ & $\tablehighlight{0.9872}$
& $0.4534$ & $0.3781$ & $0.5162$ & $0.4435$ & $0.3852$ & $0.3473$ \\

\normalsmall
& $0.9756$ & $0.9717$ & $0.9592$ & $0.9827$ & $0.9804$ & $0.9779$
& $0.4595$ & $0.3810$ & $\tablehighlight{0.5309}$ & $0.4583$ & $0.3894$ & $0.3405$ \\

\bottomrule
\end{tabular}}
\end{table}

\section{Discussion and Future Work}\vspace{-2pt}
While we observe that masked diffusion models with learned schedules lead to lower validation losses, we do not observe significant gains in downstream data fidelity metrics. A possible reason for this observation is that learned schedulers increase the loss variance during training. 
For example, even in the univariate case where loss is invariant to the noise schedule, \citet{shi2024simplified} observe that the linear schedule leads to best likelihood performance and attribute it to low loss variance from linear schedule. Since our schedules are learned and different for each column, it might be possible that loss variance is high. Future work should focus on how these effects can be mitigated for multivariate schedules. An interesting open question is whether masked diffusion with learned schedules can discover independencies and structures in tabular data.

\begin{ack}
We acknowledge the support of the SBI Foundation Hub for Data Science \& Analytics at the Indian Institute of Technology Bombay for providing financial support and infrastructure for conducting the research presented in this paper.
\end{ack}

\bibliography{iclr2026_conference}
\bibliographystyle{iclr2026_conference}

\newpage
\appendix

\section*{\centering Masked Diffusion Models are Secretly Learned-Order Autoregressive Models (Appendix)}

\section{Related Work}

Diffusion models have transformed generative modeling \citep{sohl-dickstein15, ho2022classifierfreediffusionguidance, dhariwal2021diffusion, ddpm, ddim, vdm, edm}, and recent work has adapted them to discrete domains such as language, graphs, and proteins \citep{d3pm, vignac2023digress, gat_discrete_2024, stark_dirichlet_2024, shaul_flow_2024, boget_simple_2025, han_discrete_2025, zhao_informed_2025, yu_discrete_2025}. Discrete diffusion has been applied to diverse structured data types, including categorical modeling \citep{multinominal_diffusion, continuous_diffusion}, text \citep{lou_discrete_2024, sahoo_diffusion_2025, amin_why_2025}, and multimodal setups \citep{yu_dimple_2025, shi_muddit_2025, campbell_generative_2024}. Recent work by \cite{sahoo_diffusion_2024} applies multivariate in continuous diffusion setting. We instead focus on discrete diffusion processes elucidating a direct connections to decoding orders. \cite{shi2024simplified} also introduces GenMD4, where instead of token position, token values dictate the noise schedules which results in a significantly complicated objective. We instead show that a simple extension to the MDM objective works for token position dependent schedules.

The role of ordering and masking schedules in discrete diffusion has only recently received attention. Recent studies investigate inference heuristic or learned sampler for improved sample quality \citep{peng_path_2025, kim_train_2025}. Findings suggest that ordering choices significantly influence model inference quality. \cite{wang_learning-order_2025} do not proceed from a diffusion setup, rather models decoding orders as a latent variable. Our work instead approaches this from discrete point of view, connecting them to learned-order autoregressive model.

\section{Proofs}
\cdftranstime*
\pfref{prop:cdf-transition-time}: Consider a discretization of time interval $[0,1]$ into $T$ steps. We define $s(i) = \frac{i-1}{T}$ and $t(i) = \frac{i}{T}$, Let $E_i$ be the event that the token transition to a non-mask state for the \textit{first time} when transitioning from $t(i)$ to $s(i)$
We denote the event that it transition in the $i$-th with $E'_i$. We know that from the definition of reverse process, $P(\neg E'_i) = 1 - \frac{\asi - \ati}{1 - \ati} = \frac{1 - \asi}{1 - \ati}$

\begin{align}\label{eqn:proof-cdf}
    P(E_i) &= P(E'_i)\prod_{i<j} P(\neg E'_j) \\
           &= (1 -  P(\neg E'_i))\prod_{i<j} P(\neg E'_j) \\
           &= \prod_{i<j} P(\neg E'_j) - \prod_{i\leq j} P(\neg E'_j) \\
           &= \prod_{i<j} \frac{1 - \asj}{1 - \atj} - \prod_{i\leq j} \frac{1 - \asi}{1 - \ati} \\
        &=\frac{1 - \alpha_{s(i+1)}}{1 - \alpha_{t(T)}} - \frac{1 - \asi}{1 - \alpha_{t(T)}}\\
        &=\frac{1 - \alpha_{t(i)}}{1 - 0} - \frac{1 - \asi}{1 - 0}\\
        &=\asi - \ati
\end{align}

\begin{align}
    P(\tau \leq t) = P\left(\bigcup_{i=0}^{i^*(t)} E_i\right)
\end{align}
$i^*(t) = \argmax_i s(i) \leq t < t(i)$. Since $E_i$'s are disjoint events, we have,
\begin{align}
    P(\tau \leq t) &= \sum_{i=0}^{i^*(t)} P(E_i) \\
    &= \sum_{i=1}^{i^*(t)} \asi - \ati \\
    &= 1 - \alpha_{t(i^*)}
\end{align}
As $T \rightarrow \infty$, $t(i^*) \rightarrow t$, so we get, 
\begin{align}
    P(\tau \leq t) = 1 - \alpha_{t}
\end{align}

\elboloarm*

\pfref{prop:elbo-lo-arm}
We will analyse the expression in~\ref{eqn:elbo-trajectory} 

\begin{align}\label{eqn:elbo-trajectory-simp}
    \lossfinal = \mathbb{E}_q\!\left[
    - \log p(\x_T)
    - \sum_{t \ge 1}
        (\log p_\theta(\x_{s(i)} \mid \x_{t(i)})
        - \log q(\x_{t(i)}\mid \x_{s(i)}))
\right]
\end{align}

Crucial thing to note is that this expectation is taken over all possible trajectories as shown in \pref{fig:forwardprocps}
With $\bts = \frac{1 - \as}{1- \at}$, we define the $p_\theta(\x_{s(i)} \mid \x_{t(i)})$ and $q(\x_{t(i)}\mid \x_{s(i)})$ in the table below
\begin{center}
    \begin{tabular}{cccc}
\toprule
$s$ & $t$ & $p$ & $q$ \\
\midrule
$\m$ & $\m$ & $1$ & $\bts$ \\
$\m$ & $\x$ & $0$ & $0$ \\
$\x$ & $\x$ & $\ats$ & $1$ \\
$\x$ & $\m$ & $(1 - \ats) \cdot \mu^x(\m, t^*)$ & $1 - \bts$ \\
\bottomrule
\end{tabular}
\end{center}
where $t^*$ is the time token transitions in the reverse process. We can safely ignore the case $\m, \x$ because it happens with probability $0$. Consider a discretization of time interval $[0,1]$ into $T$ steps. We define $s(i) = \frac{i-1}{T}$ and $t(i) = \frac{i}{T}$,

\begin{align}\label{eqn:elbo-trajectory-terms}
    \log p_\theta(\x_{s(i)} \mid \x_{t(i)})
        - \log q(\x_{t(i)}\mid \x_{s(i)}) = \begin{cases}
            -\log(\bts(i)) \quad  i < i^* \\
            \log (1 - \ats(i)) + \log \mu^x(\m, t(i^*))  - \log(1 - \bts(i)) \quad  i = i^* \\
            \log \ats(i) \quad  i > i^* \\
        \end{cases}
\end{align}
$i^*$ refers to the discrete step where the transition happens. Summing it up, it reduces to telescopic sums and the only term which remains is $\log \mu^\x(\m, t(i^*))$. So, rewriting~\pref{eqn:elbo-trajectory-simp}, we have,

\begin{align}\label{eqn:elbo-trajectory-mu}
    \lossfinal = \mathbb{E}_q\!\left[\log \mu^x(\m, t(i^*))\right]
\end{align}

For a sequence of token $\x^{1:L}$, it will have corresponding transition times $\{\t^*_{\ell}\}_{\ell=1}^L$. In continuous time limit, two tokens do not transition at the same time ($p(\t^*_{i} =\t^*_{j}. i\neq j ) = 0$). For given transition times we define a map $\pi$ such that $t^*_{\pi(1)} < t^*_{\pi(2)} < \cdots t^*_{\pi(L)} $, it is easy to see that $\pi$ is a permutation. Corresponding $\x_t$'s at these transition time will have the corresponding masks (specifically at $t^*_{\pi(i)}$, all the indices corresponding to $\pi(\leq i)$, would be masked, and rest will be same as $\x_0$)

\begin{align}\label{eqn:elbo-trajectory-seq}
    \lossfinal = \mathbb{E}_q\!\left[\sum_\ell\log \mu^{\x_{\pi(\ell)}}(\x_{t_\pi(\ell)^*}, t_{\pi(\ell)}^*))\right]
\end{align}

Given an order $\pi$, we can define joint distribution over $\{t^*_\ell\}$, using~\pref{prop:cdf-transition-time}. 
\begin{align}\label{eqn:order-distribution}
    P(\pi) = \int_{\Omega_\pi} \left(\prod_\ell -\datl\right) 
\end{align}
where $\Omega_\pi = \{t: t_{\pi(1)} < t_{\pi(2)} < \cdots < t_{\pi(L)} \}$

\begin{align}
    p(t^*_{1:L} \mid \pi) = \left(\frac{\prod_\ell -\alpha'_{t^*,\ell}}{P(\pi)}\right)I[t^* \in\Omega_\pi] 
\end{align}

We can collect all the trajectories which result in permutation $\pi$ and thus we arrive at the result,
    \begin{align}
        \lossfinalseq &= -\mathbb{E}_{\pi}\left[\sum_i\mathbb{E}_{t_{\pi(i)}|\pi}\left[\log\bmu(\x_{\pi(i)}|\x_{\pi(<i)},t_{\pi(i)}; \theta)\right]\right] 
    \end{align}
\section{Experiments Details and Additional Results}\label{app:exptdeets}

\subsection{Performance Metrics and Baseline Comparisons}\label{app:addperf}
For the tables below, results for \rloosmall~and\normalsmall~are averaged over 10 runs. The results of other baselines are taken from~\citet{shi_tabdiff_2025}.

\begin{table}[H]
\centering\caption{\densitytrend}\vspace{6pt}
\resizebox{\textwidth}{!}{\begin{tabular}{lrrrrrr}
\toprule
Method & \ccol{Adult} & \ccol{Default} & \ccol{Beijing} & \ccol{Shoppers} & \ccol{Magic} & \ccol{News} \\
\midrule
STaSy & $0.8549${\tiny$\pm0.0025$} & $0.9404${\tiny$\pm0.0026$} & $0.9151${\tiny$\pm0.0015$} & $0.9339${\tiny$\pm0.0053$} & $0.9200${\tiny$\pm0.0010$} & $0.9693${\tiny$\pm0.0004$} \\
CoDi & $0.7751${\tiny$\pm0.0008$} & $0.3159${\tiny$\pm0.0005$} & $0.8222${\tiny$\pm0.0011$} & $0.9347${\tiny$\pm0.0025$} & $0.9293${\tiny$\pm0.0015$} & $0.8890${\tiny$\pm0.0001$} \\
TabDDPM & $0.9699${\tiny$\pm0.0025$} & $0.9511${\tiny$\pm0.0010$} & $0.9339${\tiny$\pm0.0016$} & $0.9830${\tiny$\pm0.0022$} & $0.9729${\tiny$\pm0.0009$} & $0.8684${\tiny$\pm0.0011$} \\
TabSyn & $0.9807${\tiny$\pm0.0007$} & $0.9719${\tiny$\pm0.0048$} & $0.9787${\tiny$\pm0.0010$} & $0.9912${\tiny$\pm0.0018$} & $0.9687${\tiny$\pm0.0034$} & $0.9848${\tiny$\pm0.0003$} \\
TabDiff & \tablehighlight{$0.9851${\tiny$\pm0.0016$}} & \tablehighlight{$0.9745${\tiny$\pm0.0075$}} & \tablehighlight{$0.9826${\tiny$\pm0.0008$}} & \tablehighlight{$0.9924${\tiny$\pm0.0012$}} & \tablehighlight{$0.9741${\tiny$\pm0.0015$}} & \tablehighlight{$0.9872${\tiny$\pm0.0004$}} \\
\midrule\rloosmall & $0.9778${\tiny$\pm0.0023$} & $0.9705${\tiny$\pm0.0028$} & $0.9743${\tiny$\pm0.0038$} & $0.9710${\tiny$\pm0.0038$} & $0.9628${\tiny$\pm0.0070$} & $0.9781${\tiny$\pm0.0017$} \\
\normalsmall & $0.9747${\tiny$\pm0.0020$} & $0.9642${\tiny$\pm0.0016$} & $0.9773${\tiny$\pm0.0034$} & $0.9703${\tiny$\pm0.0027$} & $0.9724${\tiny$\pm0.0026$} & $0.9766${\tiny$\pm0.0016$} \\
\bottomrule
\end{tabular}
}
\end{table}

\begin{table}[H]
\centering\caption{\densityshape}\vspace{6pt}
\resizebox{\textwidth}{!}{\begin{tabular}{lrrrrrr}
\toprule
Method & \ccol{Adult} & \ccol{Default} & \ccol{Beijing} & \ccol{Shoppers} & \ccol{Magic} & \ccol{News} \\
\midrule
STaSy & $0.8871${\tiny$\pm0.0006$} & $0.9423${\tiny$\pm0.0006$} & $0.9063${\tiny$\pm0.0009$} & $0.9371${\tiny$\pm0.0013$} & $0.9329${\tiny$\pm0.0003$} & $0.9311${\tiny$\pm0.0003$} \\
CoDi & $0.7862${\tiny$\pm0.0006$} & $0.8423${\tiny$\pm0.0007$} & $0.6816${\tiny$\pm0.0005$} & $0.8844${\tiny$\pm0.0026$} & $0.8306${\tiny$\pm0.0002$} & $0.6773${\tiny$\pm0.0004$} \\
TabDDPM & $0.9825${\tiny$\pm0.0003$} & $0.9843${\tiny$\pm0.0008$} & $0.9728${\tiny$\pm0.0013$} & $0.9899${\tiny$\pm0.0009$} & $0.9870${\tiny$\pm0.0003$} & $0.2125${\tiny$\pm0.0001$} \\
TabSyn & $0.9919${\tiny$\pm0.0005$} & \tablehighlight{$0.9899${\tiny$\pm0.0008$}} & $0.9856${\tiny$\pm0.0007$} & $0.9897${\tiny$\pm0.0014$} & $0.9874${\tiny$\pm0.0005$} & \tablehighlight{$0.9794${\tiny$\pm0.0004$}} \\
TabDiff & \tablehighlight{$0.9937${\tiny$\pm0.0005$}} & $0.9876${\tiny$\pm0.0007$} & \tablehighlight{$0.9872${\tiny$\pm0.0009$}} & \tablehighlight{$0.9922${\tiny$\pm0.0008$}} & \tablehighlight{$0.9897${\tiny$\pm0.0005$}} & $0.9765${\tiny$\pm0.0003$} \\
\midrule\rloosmall & $0.9878${\tiny$\pm0.0016$} & $0.9830${\tiny$\pm0.0007$} & $0.9841${\tiny$\pm0.0009$} & $0.9832${\tiny$\pm0.0018$} & $0.9750${\tiny$\pm0.0044$} & $0.9787${\tiny$\pm0.0004$} \\
\normalsmall & $0.9851${\tiny$\pm0.0015$} & $0.9787${\tiny$\pm0.0010$} & $0.9866${\tiny$\pm0.0006$} & $0.9829${\tiny$\pm0.0011$} & $0.9822${\tiny$\pm0.0018$} & $0.9785${\tiny$\pm0.0004$} \\
\bottomrule
\end{tabular}
}
\end{table}

\begin{table}[H]
\centering\caption{\alphaprecision}\vspace{6pt}
\resizebox{\textwidth}{!}{\begin{tabular}{lrrrrrr}
\toprule
Method & \ccol{Adult} & \ccol{Default} & \ccol{Beijing} & \ccol{Shoppers} & \ccol{Magic} & \ccol{News} \\
\midrule
STaSy & $0.8287${\tiny$\pm0.0026$} & $0.9048${\tiny$\pm0.0011$} & $0.8965${\tiny$\pm0.0025$} & $0.8656${\tiny$\pm0.0019$} & $0.8916${\tiny$\pm0.0012$} & $0.9476${\tiny$\pm0.0033$} \\
CoDi & $0.7758${\tiny$\pm0.0045$} & $0.8238${\tiny$\pm0.0015$} & $0.9495${\tiny$\pm0.0035$} & $0.8501${\tiny$\pm0.0036$} & \tablehighlight{$0.9813${\tiny$\pm0.0038$}} & $0.8715${\tiny$\pm0.0012$} \\
TabDDPM & $0.9636${\tiny$\pm0.0020$} & $0.9759${\tiny$\pm0.0036$} & $0.8855${\tiny$\pm0.0068$} & $0.9859${\tiny$\pm0.0017$} & $0.9793${\tiny$\pm0.0030$} & $0.0000${\tiny$\pm0.0000$} \\
TabSyn & \tablehighlight{$0.9939${\tiny$\pm0.0018$}} & \tablehighlight{$0.9865${\tiny$\pm0.0023$}} & $0.9836${\tiny$\pm0.0052$} & $0.9942${\tiny$\pm0.0028$} & $0.9751${\tiny$\pm0.0024$} & $0.9505${\tiny$\pm0.0030$} \\
TabDiff & $0.9902${\tiny$\pm0.0020$} & $0.9849${\tiny$\pm0.0028$} & \tablehighlight{$0.9911${\tiny$\pm0.0034$}} & \tablehighlight{$0.9947${\tiny$\pm0.0021$}} & $0.9806${\tiny$\pm0.0024$} & $0.9736${\tiny$\pm0.0017$} \\
\midrule\rloosmall & $0.9909${\tiny$\pm0.0028$} & $0.9839${\tiny$\pm0.0019$} & $0.9744${\tiny$\pm0.0034$} & $0.9818${\tiny$\pm0.0048$} & $0.9623${\tiny$\pm0.0034$} & \tablehighlight{$0.9872${\tiny$\pm0.0012$}} \\
\normalsmall & $0.9756${\tiny$\pm0.0032$} & $0.9717${\tiny$\pm0.0030$} & $0.9592${\tiny$\pm0.0022$} & $0.9827${\tiny$\pm0.0033$} & $0.9804${\tiny$\pm0.0023$} & $0.9779${\tiny$\pm0.0016$} \\
\bottomrule
\end{tabular}
}
\end{table}\vspace{-1em}

\begin{table}[H]
\centering\caption{\betarecall}\vspace{6pt}
\resizebox{\textwidth}{!}{\begin{tabular}{lrrrrrr}
\toprule
Method & \ccol{Adult} & \ccol{Default} & \ccol{Beijing} & \ccol{Shoppers} & \ccol{Magic} & \ccol{News} \\
\midrule
STaSy & $0.2921${\tiny$\pm0.0034$} & $0.3931${\tiny$\pm0.0039$} & $0.3724${\tiny$\pm0.0045$} & \tablehighlight{$0.5397${\tiny$\pm0.0057$}} & $0.5479${\tiny$\pm0.0018$} & $0.3942${\tiny$\pm0.0032$} \\
CoDi & $0.0920${\tiny$\pm0.0015$} & $0.1994${\tiny$\pm0.0022$} & $0.2082${\tiny$\pm0.0023$} & $0.5056${\tiny$\pm0.0031$} & $0.5219${\tiny$\pm0.0012$} & $0.3440${\tiny$\pm0.0031$} \\
TabDDPM & $0.4705${\tiny$\pm0.0025$} & $0.4783${\tiny$\pm0.0035$} & $0.4779${\tiny$\pm0.0025$} & $0.4846${\tiny$\pm0.0042$} & $0.5692${\tiny$\pm0.0013$} & $0.0000${\tiny$\pm0.0000$} \\
TabSyn & $0.4792${\tiny$\pm0.0023$} & $0.4645${\tiny$\pm0.0035$} & $0.4910${\tiny$\pm0.0060$} & $0.4803${\tiny$\pm0.0050$} & $0.5915${\tiny$\pm0.0022$} & \tablehighlight{$0.4301${\tiny$\pm0.0028$}} \\
TabDiff & \tablehighlight{$0.5164${\tiny$\pm0.0020$}} & \tablehighlight{$0.5109${\tiny$\pm0.0025$}} & $0.4975${\tiny$\pm0.0064$} & $0.4801${\tiny$\pm0.0031$} & \tablehighlight{$0.5963${\tiny$\pm0.0023$}} & $0.4210${\tiny$\pm0.0032$} \\
\midrule\rloosmall & $0.4534${\tiny$\pm0.0020$} & $0.3781${\tiny$\pm0.0012$} & $0.5162${\tiny$\pm0.0027$} & $0.4435${\tiny$\pm0.0060$} & $0.3852${\tiny$\pm0.0057$} & $0.3473${\tiny$\pm0.0033$} \\
\normalsmall & $0.4595${\tiny$\pm0.0022$} & $0.3810${\tiny$\pm0.0030$} & \tablehighlight{$0.5309${\tiny$\pm0.0031$}} & $0.4583${\tiny$\pm0.0067$} & $0.3894${\tiny$\pm0.0042$} & $0.3405${\tiny$\pm0.0022$} \\
\bottomrule
\end{tabular}
}
\end{table}

\subsection{Visualisation of Learned Schedules}\label{app:learnedschedule}

\begin{figure}[H]
    \centering
    \caption{Schedules learned by TabDiff, reproduced from original implementation}
    \includegraphics[width=\linewidth]{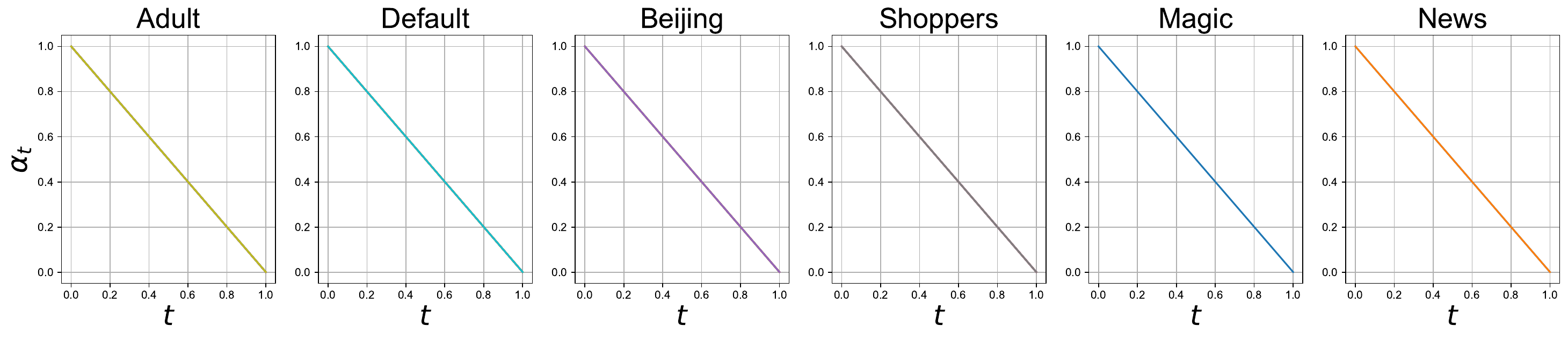}
    \label{fig:noise_tabdiff}
\end{figure}\vspace{-2em}

\begin{figure}[H]
    \centering
    \caption{Tabdiff fails to learn schedules even with random initialization}
    \includegraphics[width=\linewidth]{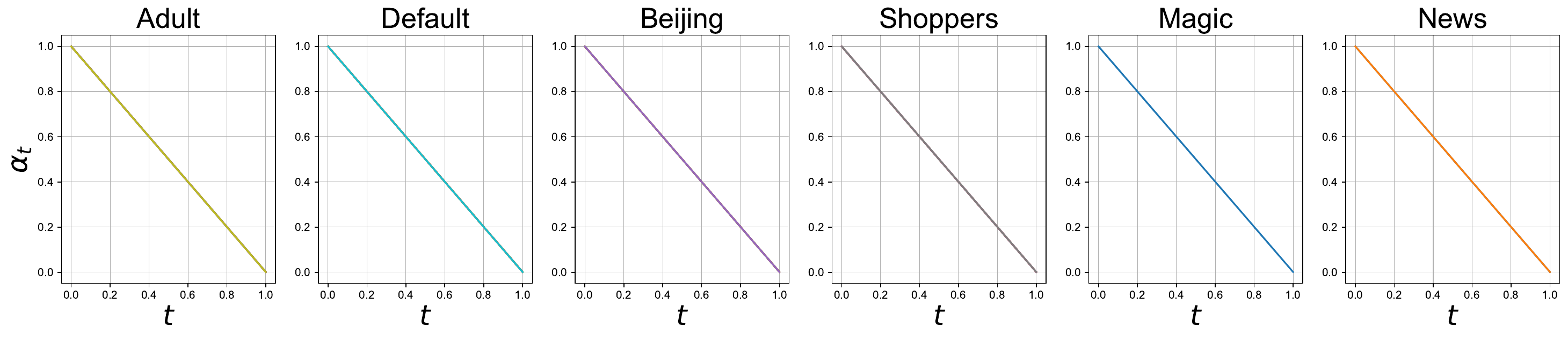}
    \label{fig:noise_tabdiff_rand}
\end{figure}\vspace{-2em}

\begin{figure}[H]
    \centering
    \caption{Schedules learned by our implementation of \rloosmall. Note that our implementation uses masked diffusion for all feature columns and not just for categorical features.}
    \includegraphics[width=\linewidth]{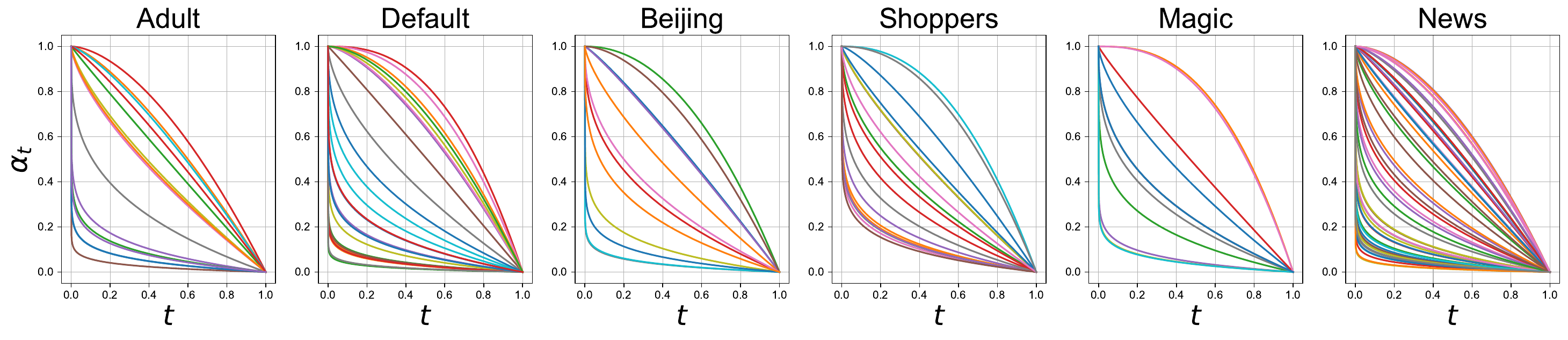}
    \label{fig:noise_md4}
\end{figure}\vspace{-2em}

TabDiff makes use of a hybrid setup with masked diffusion for categorical columns with column wise learnable schedules. From our reproduction, we see that it fails to learn the schedules
In~\pref{fig:noise_tabdiff}, we train the models from the implementation provided by the authors\footnote{https://github.com/MinkaiXu/TabDiff} and plot the schedules learned at the end of training. In~\pref{fig:noise_tabdiff_rand}, we attempt to break symmetry by randomizing the initialization of the noise schedule, which unfortunately leads to the same result. For comparison, we include the schedules learned by our \rloosmall~models in~\pref{fig:noise_md4}.

\end{document}